\documentclass[conference]{IEEEtran}
\IEEEoverridecommandlockouts
\usepackage{cite}
\usepackage{amsmath,amssymb,amsfonts}
\usepackage{algorithmic}
\usepackage{graphicx}
\usepackage{caption}
\usepackage{subcaption}
\usepackage{textcomp}
\usepackage{xcolor}
\usepackage[ruled,vlined]{algorithm2e}
\usepackage{caption}
\def\BibTeX{{\rm B\kern-.05em{\sc i\kern-.025em b}\kern-.08em
    T\kern-.1667em\lower.7ex\hbox{E}\kern-.125emX}}
    
\makeatletter
\newcommand{\newlineauthors}{%
  \end{@IEEEauthorhalign}\hfill\mbox{}\par
  \mbox{}\hfill\begin{@IEEEauthorhalign}
}
\makeatother    
    
\begin{document}

\title{Towards a Multi-purpose Robotic Nursing Assistant\\
{\footnotesize }
\thanks{This work was supported by the National Science Foundation (NSF) grant IIP 1719031. }
}

\author{
\IEEEauthorblockN{Krishna Chaitanya Kodur}
\IEEEauthorblockA{\textit{Computer Science \& Engineering Dept.} \\
\textit{University of Texas at Arlington}\\
Arlington,TX, USA \\
kck8298@mavs.uta.edu}
\and
\IEEEauthorblockN{Kaustubh Rajpathak}
\IEEEauthorblockA{\textit{Computer Science \& Engineering Dept.} \\
\textit{University of Texas at Arlington}\\
Arlington,TX, USA \\
kaustubh.rajpathak@mavs.uta.edu}
\and
\IEEEauthorblockN{Akilesh Rajavenkatanarayanan}
\IEEEauthorblockA{\textit{Computer Science \& Engineering Dept.} \\
\textit{University of Texas at Arlington}\\
Arlington,TX, USA \\
akilesh.rajavenkatanarayanan@mavs.uta.edu}
\newlineauthors
\IEEEauthorblockN{Maria Kyrarini}
\IEEEauthorblockA{\textit{Computer Science \& Engineering Dept.} \\
\textit{University of Texas at Arlington}\\
Arlington,TX, USA \\
maria.kyrarini@uta.edu}
\and
\IEEEauthorblockN{Fillia Makedon}
\IEEEauthorblockA{\textit{Computer Science \& Engineering Dept.} \\
\textit{University of Texas at Arlington}\\
Arlington,TX, USA \\
makedon@uta.edu}
}

\maketitle

\begin{abstract}
Robotic nursing aid is one of the heavily researched areas in robotics nowadays. Several robotic assistants exist that only focus on a specific function related to nurses' assistance or functions related to patient aid. There is a need for a unified system that not only performs tasks that would assist nurses and reduce their burden but also perform tasks that help a patient. In recent times, due to the COVID-19 pandemic, there is also an increase in the need for robotic assistants that have teleoperation capabilities to provide better protection against the virus spread. To address these requirements, we propose a novel Multi-purpose Intelligent Nurse Aid (MINA) robotic system that is capable of providing walking assistance to the patients and perform teleoperation tasks with an easy-to-use and intuitive Graphical User Interface (GUI). This paper also presents preliminary results from the walking assistant task that improves upon the current state-of-the-art methods and shows the developed GUI for teleoperation.
\end{abstract}

\begin{IEEEkeywords}
Robotic Assistant, Walking Assistance, Teleoperation, Gait Detection, Nursing Robot
\end{IEEEkeywords}

\section{Introduction}
The role of robotics in healthcare has been growing in the last decade \cite{b1}. This is due to the need to improve the quality and safety of care while controlling expenses \cite{b2}. Moreover, the shortage of nurses and healthcare personnel significantly impacts the quality of care \cite{b3,b4}. The COVID-19 pandemic demonstrated how vulnerable the healthcare workers are and at risk to be exposed to the virus \cite{b5,b21}.

Several robotic systems are introduced to assist with hospital logistics, disinfection of spaces, and patient screening \cite{b15,b16}. Robots have the potential to support caregivers in several routine tasks, while caregivers can focus on the actual patient care. For example, there is a variety of commercially available robots that are currently used in hospitals to take care of delivery tasks, such as the mobile robots TUG \cite{b6} and Relay \cite{b7}, or fetching objects, such as Moxi \cite{b8}, a mobile robot equipped with a robotic arm. Moreover, there are robots that focus on patient care, such as rehabilitation \cite{b9,b10}, walking assistance \cite{b11} or monitoring vital signs \cite{b12}. A recent review of robots in healthcare \cite{b1} makes it obvious that each robotic platform focuses on a specific care-giving task.

In a recent single-site cohort study \cite{b22}, 40 patients interacted with a mobile robotic system, which was controlled by a clinician in an emergency department and used to facilitate triage and telehealth tasks. After the interaction, the patients completed an assessment to measure their satisfaction. A total of 37 patients (92.5\%) reported that the interaction with the robotic system was satisfactory and 33 patients (82.5\%) reported that their experience of a mobile robotic system facilitated the interview was as satisfactory as an in-person interview with a doctor. These findings are very promising for the acceptance of robotic telehealth. 

In this work, we propose a solution designed for a multi-purpose robotic platform that has a high potential to perform several tasks in healthcare. The MINA robotic system is a collaborative robot that fetches objects (as shown in \cite{b29}), assists patients with walking, and is teleoperated by nurses in an easy and intuitive manner. This work aims to assist and protect nurses with a general-purpose robot that can be employed for many applications. The main contributions of this work are the following: (a) an improved leg detection model using laser scanner data during walking assistance tasks, (b) a multimodal approach for gait detection, and (c) an easy-to-use GUI that enables teleoperation.

The paper is organized as follows; Section 2 discusses the related work, Section 3 presents the proposed multi-purpose robotic system MINA, and Section 4 presents the preliminary results. Finally, Section 5 concludes the work and discusses future steps.

\section{Related Work}
In recent years, several robotic platforms have been developed and deployed in healthcare. There are robotic platforms that assist with rehabilitation  \cite{b9, b10}, walking \cite{b13}, and surgery \cite{b14}. During the COVID-19 pandemic, robots took over logistic tasks in hospitals, patient screening, and disinfecting tasks \cite{b15, b16}. Most of the robotic platforms are designed in an application-oriented fashion, requiring stakeholders to acquire multiple robotic platforms to assist with different tasks. 

Abubakar et al. \cite{b23} proposed an Adaptive Robotic Nursing Assistant (ARNA) robot, which is a custom-made robot by the University of Louisville in Kentucky and consists of an omnidirectional mobile base with an instrumented handlebar, and a 7-Degrees of Freedom (DoFs) robotic arm. ARNA has two primary tasks, a patient monitoring task, wherein the robot monitors a patient and responds to remote commands, and a patient walking task, wherein the patient controls the robot's motion while walking behind it. A cohort trial with 24 human subjects was conducted to evaluate ARNA, and the results of this preliminary user study indicate good usefulness and ease to use for the two primary robotic tasks.

However, in other industries, such as warehouses and manufacturing,  multi-purpose robotic systems research has advanced. An example is the Mobile Collaborative Robotic Assistant (MOCA) \cite{b17}, conceived and realized by the Italian Institute of Technology (IIT), which consists of a commercially available omnidirectional mobile base (3-DoFs) and a commercially available robotic arm (7-DoFs) equipped with a SoftHand as end-effector. MOCA has been employed in several tasks in warehouses, such as assisting humans with co-manipulation of objects \cite{b18}, transportation and positioning of a pallet jack \cite{b19}, and teleoperation \cite{b17}.  

Inspired by the MOCA system, the MINA system incorporates many functionalities present in application-oriented systems and aims at combining them in a user-friendly manner to assist with several healthcare tasks.

\section{Multi-purpose Robotic System}

\subsection{Hardware}
\begin{figure}[h]
\centering
\includegraphics[scale = 0.4]{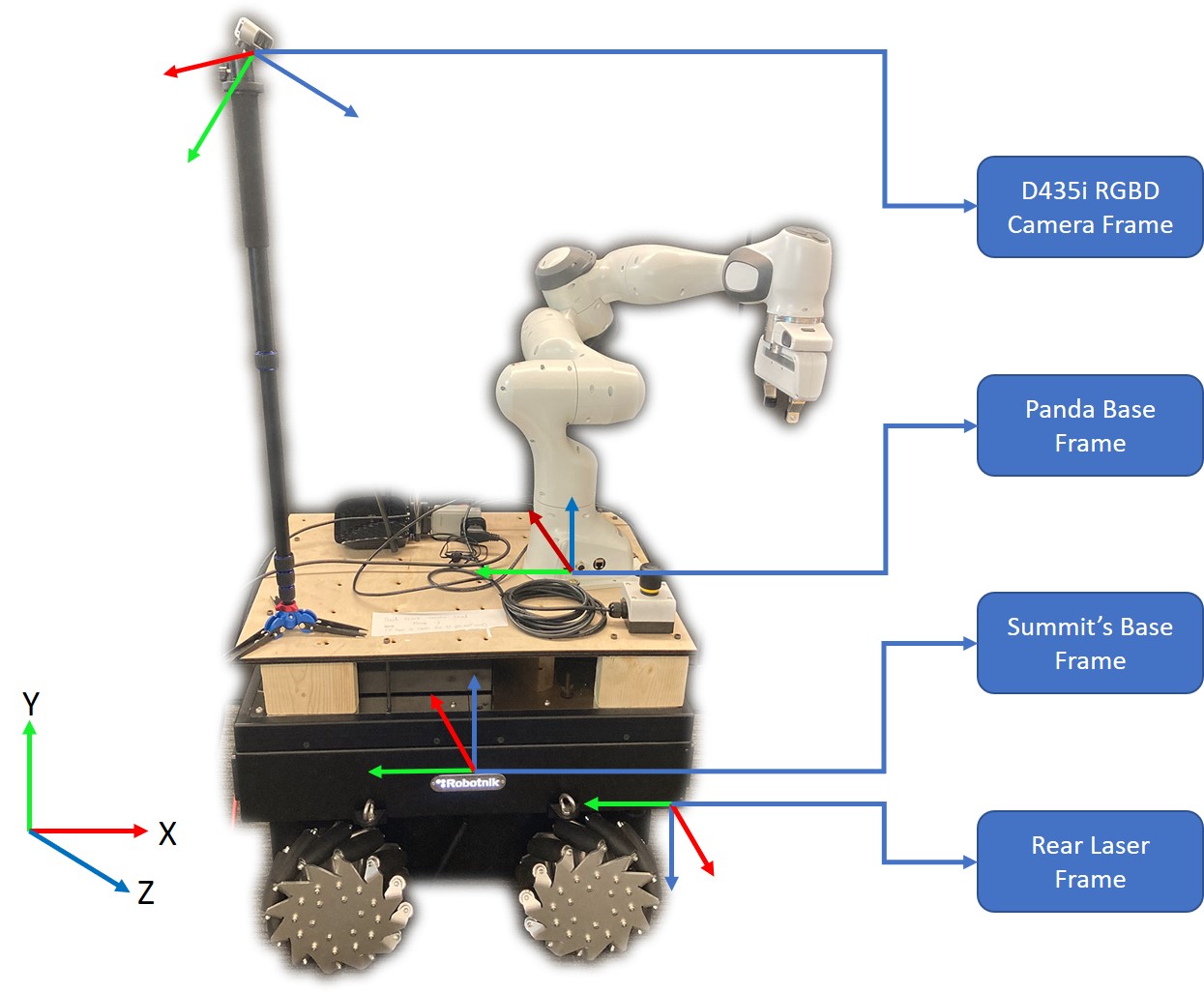}
\caption{The MINA Robotic System and the coordinate frames of its components.} 
\label{fig:system_architecture}
\end{figure}

The MINA system consists of the following components; (a) the Robotnik SUMMIT-XL STEEL, which is a 3-DoFs omnidirectional mobile base and it is equipped with two UST 20LX Hokuyo laser scanners, (b) the Franka Emika Panda, which is a 7-DoFs robotic arm equipped with a 2-finger gripper, and (c) the Intel RealSense D435i camera, which integrates an RGB camera and a Depth sensor. Fig. \ref{fig:system_architecture} shows the MINA system and the coordinate frames of its components.  

\subsection{Walking Assistance Task} \label{sec:WAT}
One of MINA's primary purposes is to assist a user with walking and, while doing so, the system calculates the gait parameters that is a valuable information of the patient's progress. An overview of the walking assistance framework is shown in Fig. \ref{walking}. The camera captures the user's mobility and the OpenPose algorithm \cite{b24} detects the body keypoints, which are then used to calculate the gait parameters. The OpenPose provides body keypoints as pixel coordinates. These pixel coordinates are then de-projected to get the 3D coordinates of each keypoints.

In our setup, the robotic arm is used as a support while walking. During this task, the user's lower body is occluded by the mobile base, which makes it difficult to detect keypoints related to the legs and ankles for gait estimation. To address this issue, we make use of the rear laser scanner that is available on the mobile base. 
\begin{center}
\includegraphics[scale=0.37]{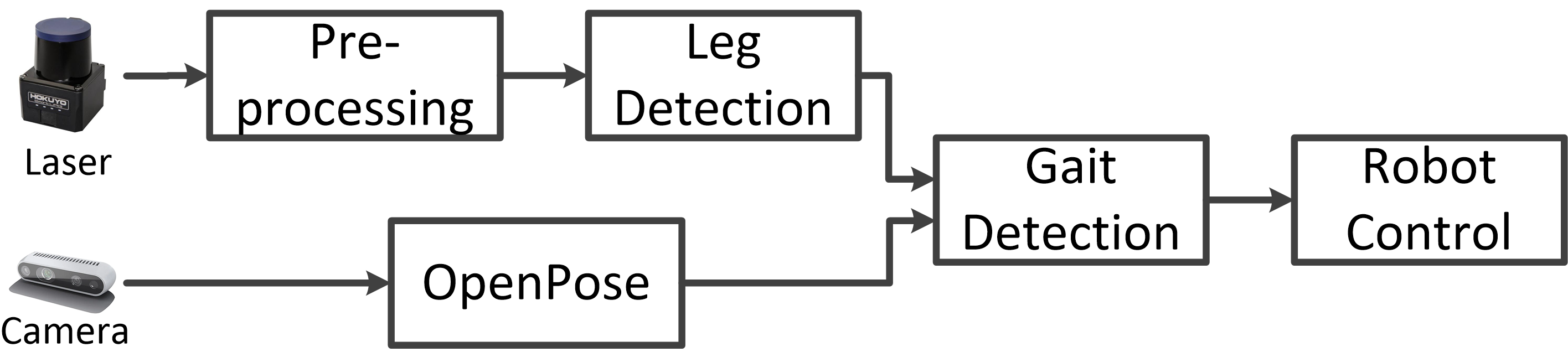}
\captionof{figure}{Overview of the Walking Assistance Framework}
\label{walking}
\end{center}

To detect the legs from the laser scan, we generate an occupancy grid which is used as input to a Convolutional Neural Network model (CNN) to segment the legs. An occupancy grid is a binary grayscale image of size 256x256 generated to detect the leg's keypoints from the laser scanner using algorithm \ref{algo:occupany_grid}.

\begin{figure}[h]
\centering
\includegraphics[scale = 0.58]{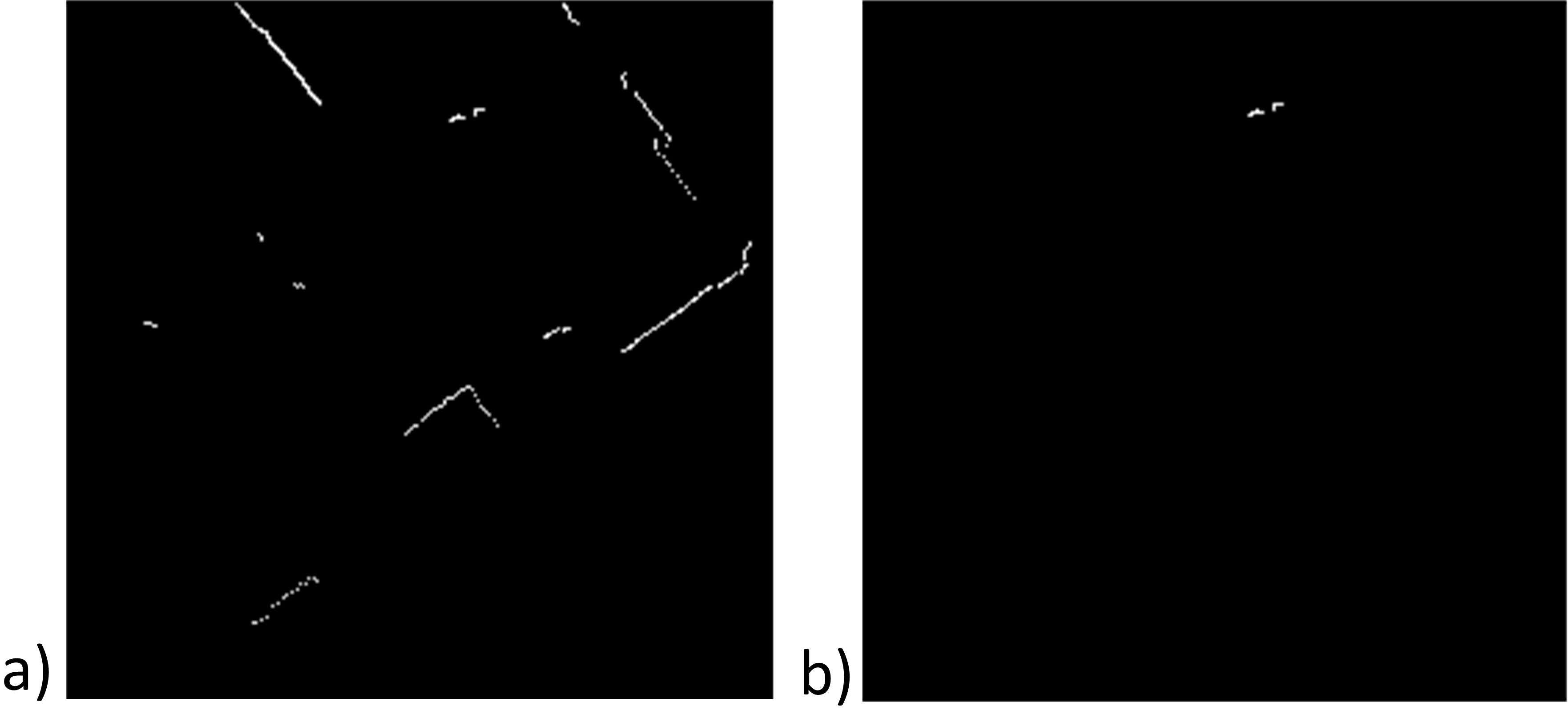}
\caption{(a) Occupancy grid obtained from laser scanner, (b) Segmented legs (two white blobs) from the occupancy grid image }
\label{fig:occu_grid}
\end{figure}

\begin{algorithm}
\SetAlgoLined
\KwResult{occupancy\_grid (256x256 binary grayscale image)}

 matrix\_length = 256;
 
 \textit{l} = matrix\_length/2;
 
 occupancy\_grid = zeros(256, 256);
 
 \For{angle i, obstacle\_distance from range of angles scanned by laser}{
  // Calculate x and y which are the real world coordinates in laser frame  
  
  x = obstacle\_distance $\times$ cos(\textit{i}); 
  
  y = obstacle\_distance $\times$ sin(\textit{i});  
  
  $pixel_x$ = x $\times$ 100 + \textit{l};  
  
  $pixel_y$ = y $\times$ 100 + \textit{l};  
  
  \If{pixel\_x and pixel\_y between 0 and matrix\_length}{  
  occupancy\_grid[$pixel_x$][$pixel_y$] = 255;  
  }
 }
\caption{Occupancy Grid Generation: Laser Ranges to Pixels}
\label{algo:occupany_grid}
\end{algorithm}
This image is then fed into a U-Net model\cite{b28}, a commonly used CNN for segmentation tasks. In this scenario, we try to segment the user's legs from the environmental clutter. For instance, Fig. \ref{fig:occu_grid}a shows the occupancy grid, and Fig. \ref{fig:occu_grid}b shows a de-cluttered image showing only the detected legs.

As a baseline model for leg detection, we use the U-Net model proposed by \cite{b25}. For our system, we propose the MINA model that improves upon the model proposed by \cite{b25} by retraining the U-Net model with dataset available at \cite{b26} and augmenting it with several transformations of the original dataset. We also utilize the weighted binary cross-entropy loss \cite{b27} to train the model. Once the model segments the legs, each leg's midpoint is found by calculating the respective leg's blob midpoint. This midpoint is in the form of pixel coordinates, which can be converted to 3D coordinates, as shown in (\ref{laser_pixel_to_coordinates}). 

\begin{equation} \label{laser_pixel_to_coordinates}
\begin{split}
l &= \frac{matrix\_length}{2} \\
x & = \frac{100 \times pixel_x + l}{2} \\
y & = \frac{100 \times pixel_y + l}{2} \\
z & = 0 \\
\end{split}
\end{equation}
where matrix\_length is the required image shape, pixel\_x and pixel\_y are the detected legs' midpoint in the x and y axes. Here, z is 0 because the laser scanner in use is 2D, and its scan spans only in the x-y plane. 

The 3D body keypoints are with the respect to the camera coordinate frame (C), and 3D leg keypoints from the laser sensor is in the laser coordinate frame (L). All the keypoints data need to be with respect to the same coordinate frame in order to fuse the multimodal data for gait estimation and robot control. To achieve this, each keypoint is transformed to the robot's base coordinate frame (R) using (\ref{RGBD_to_RF}) and (\ref{laser_to_RF}). In the aforementioned equations ${}^{}_{x}P$ is an 3D point in coordinate frame x where x $\in$ (C, L).  ${}^{x}_{y}M$ is a conversion matrix that converts the 3D points from frame x to frame y where y $\in$ R.

\begin{equation} \label{RGBD_to_RF}
{}^{}_{C}P \times {}^{C}_{R}M = {}^{}_{R}P
\end{equation}

\begin{equation} \label{laser_to_RF}
{}^{}_{L}P \times {}^{L}_{R}M = {}^{}_{R}P
\end{equation}

After the legs are segmented and validated, gait parameters, such as stride velocity and stride length are calculated. This stride velocity acts as input to the mobile base to set in which direction and at what speed the robot should move forward, thus automatically adjusting to the user's speed and direction.

\subsection{Teleoperation Task}
Another important task of a robotic system that assists nurses is the ability to enable nurses to care for patients that are in isolation (e.g., patients infected by an easily transmissible disease). For example, during the COVID-19 pandemic, nurses are required to gear up with personal protective equipment when they are in contact with an infected patient. Robotic teleoperation provides an easy alternative where a nurse controls the MINA system to provide a drink to the patient.

\begin{figure}[h]
\begin{center}
\includegraphics[scale=0.262]{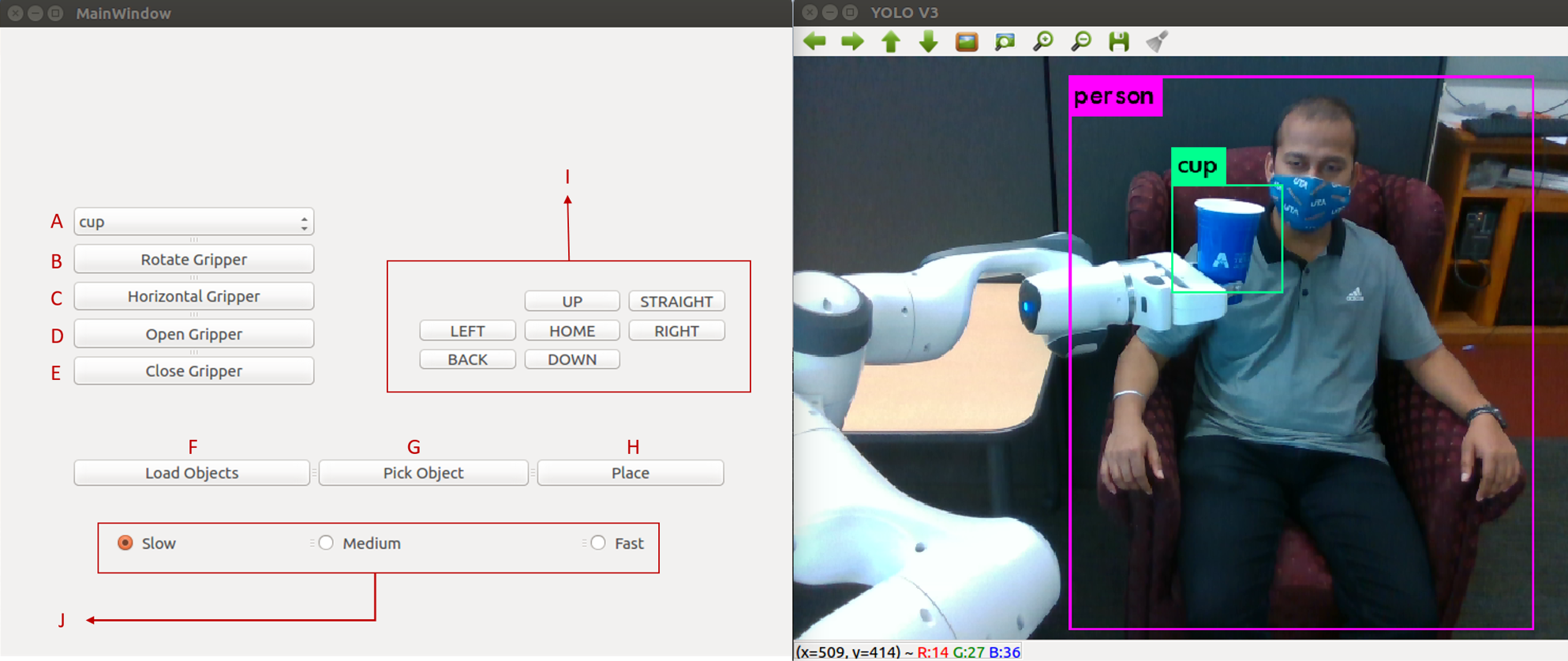}
\captionof{figure}{Screenshot of the teleoperation GUI. Right Image: YOLO Output. Left Image: Robotic arm control - (A) a drop down menu to select detected objects, (B) \& (C) orientation change of the gripper, (D) \& (E) gripper actuation/de-actuation, (F) update objects (e.g. person, cup) from the scene in (A), (G) pick the selected object, (H) place the object to a pre-defined location, (I) a cluster of motion control buttons and (J) speed selection. Mobile base control: operated by the computer keyboard using W,A,S,D keys.}
\label{minagui}
\end{center}
\end{figure}

\begin{figure}[h]
\begin{center}
\includegraphics[scale=0.05]{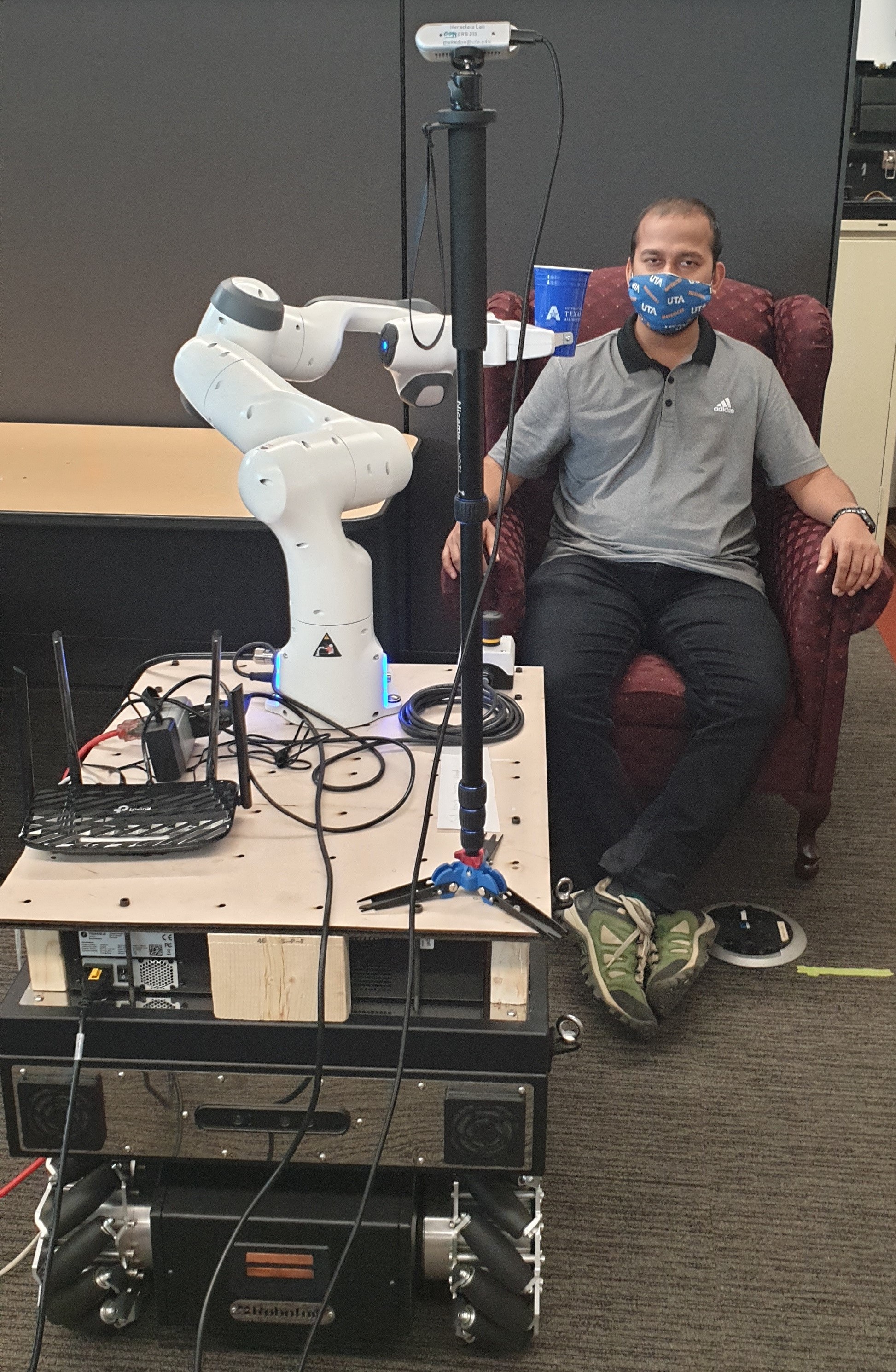}
\captionof{figure}{The MINA System in a Teleoperation Task}
\centering
\label{minademo}
\end{center}
\end{figure}

In this work, we have developed a GUI, which is easy to use and requires minimal training for the nurses. Fig. \ref{minagui} shows the GUI that consists of two parts. The first part (Fig. \ref{minagui} - Right) shows the output of object detection system YOLO (You Only Look Once)  \cite{b20}, which is the camera feed, the bounding boxes, and the name of the detected objects. The second part (Fig. \ref{minagui} - Left) has several buttons that enable the control of the robot's end-effector position in 3-dimensional space and its orientation and gripper actuation. It also enables the user to select the detected objects by YOLO and to enable an automatic pick and place of the selected objects.  Figure \ref{minademo} shows an image of the system while it is teleoperated through the GUI to provide a drink to a user.

\section{Preliminary Results} \label{sec:prelim_results}
The MINA model (see Section \ref{sec:WAT}) is compared with the baseline model proposed by Guerrero-Higueras et al. \cite{b25}. The laser scan from the MINA mobile base is converted into an occupancy grid image and is used as input for both the models, and the resulting leg segmentation results are compared. The nine locations around the robot are chosen to compare the model outputs and generate statistics. 

\begin{figure}[h]
\centering
\includegraphics[scale = 0.6]{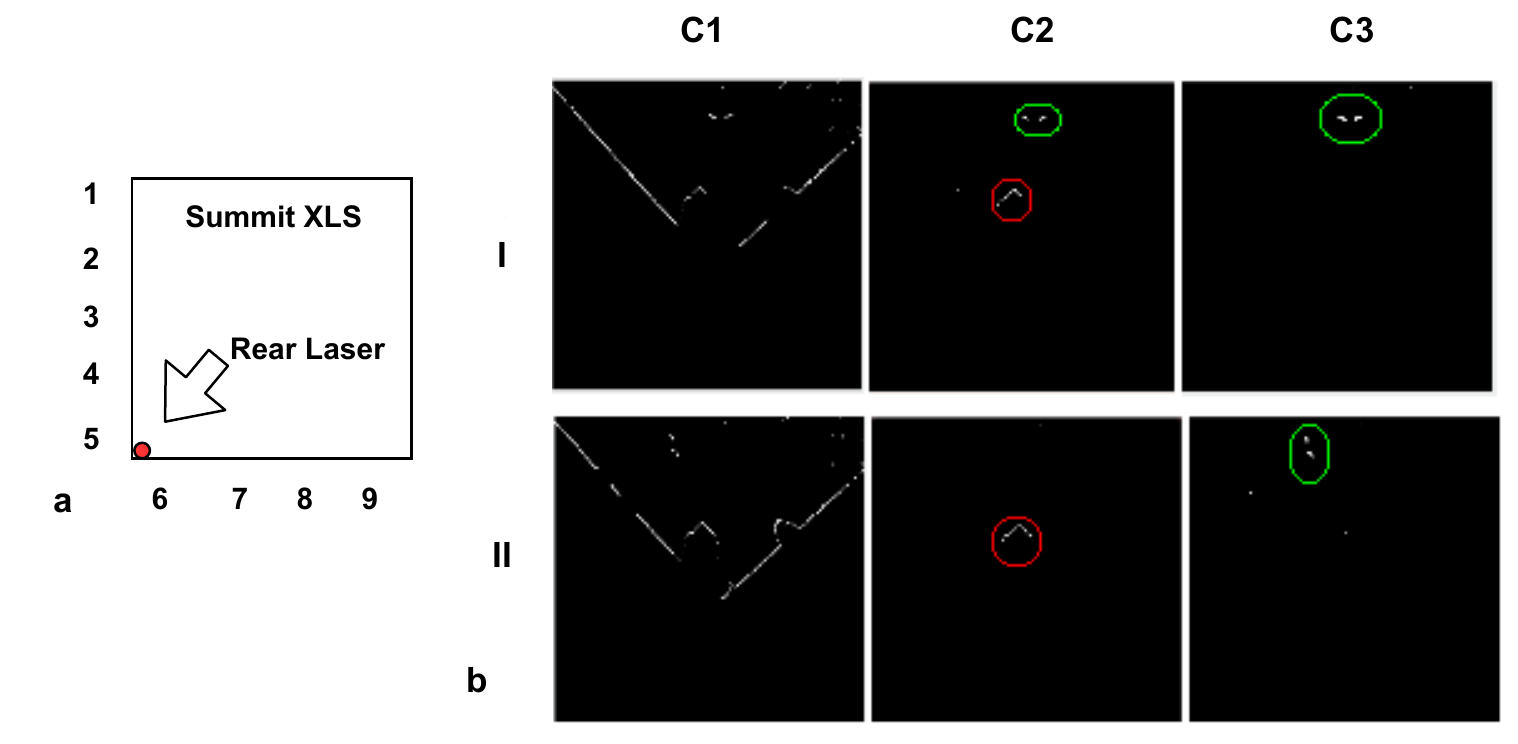}
\caption{(a) Nine different locations around the robot where a user can stand and the U-NET has to detect legs if a person stands at any location (b)  Column C1: the occupancy grid from laser, C2: segmented output from baseline, C3: segmented output from MINA model. Row I: images pertaining to Scenario -1 location 5, Row II: images pertaining to Scenario-1 location 9. The red circles show false positives. The green circles show the legs correctly segmented.}
\label{fig:comp}
\end{figure}

From table \ref{table:results}, the baseline and our MINA model are evaluated on two scenarios. One scenario is more cluttered than the other. The Scenario with more clutter has a number of small boxes on the ground, which are usually mistaken as human legs by the model. In each scenario the model is evaluated while a human stands at nine different locations as shown in Fig. \ref{fig:comp}a. Let $n_t$ be the total number of times the model is evaluated by giving different inputs by varying the locations and scenarios, $n_s$ be the total number of times when the model correctly segmented the legs, the accuracy (\textit{acc}) can be calculated using (\ref{eq:accuracy}).

\begin{equation} \label{eq:accuracy}
acc = \frac{n_s}{n_t} \times 100
\end{equation}

For the baseline model, the \textit{acc} is 77.7\%, and for our model (MINA), it is 94.4\%. Another evaluation metric is called False Positive rate (\textit{FP}); that measures if any other object except legs is segmented. Let $n_f$ be the total number of times when the model incorrectly segments other environment clutter as legs, the \textit{FP}  can be calculated using (\ref{eq:fp_rate}).

\begin{equation} \label{eq:fp_rate}
FP = \frac{n_f}{n_t} \times 100
\end{equation}

For the baseline model, the \textit{FP} is 38.8\%, and for our model MINA, it is only 5.5\%. It is clear that our model performs better than the baseline model in both \textit{acc} and \textit{FP} metrics. One factor that contributed reducing the \textit{FP} is that the U-Net is trained with data augmentations, such as random rotation. Segmentation results from two different locations are shown in Fig. \ref{fig:comp}b.  

\begin{table}[ht]
\begin{center}
\caption{Models Comparison}
\begin{tabular}{|c|c|c|c|}
\hline
\multicolumn{2}{|c|}{\textbf{Baseline Model \cite{b25}}} & \multicolumn{2}{c|}{\textbf{MINA Model (Ours)}}           \\ \hline
\textbf{Legs detected}             & \textbf{False Positives}            & \textbf{Legs detected} & \textbf{False Positives} \\ \hline
14/18                              & 7/18                                & 17/18                  & 1/18                     \\ \hline
\end{tabular}
\label{table:results}
\end{center}
\end{table} 

\section{Conclusion and Future Work}
In this work, we proposed a novel Multipurpose Intelligent Nurse Aid (MINA) collaborative robot system that is capable of assisting users with walking and is teleoperated in an easy and intuitive manner. We presented preliminary results of our leg detection algorithm that showed improved accuracy and false positive rates in comparison to the state-of-the-art leg detection algorithm. We also show the GUI developed for teleoperation of robotic manipulation that is easy to use and requires minimal training. We are currently working to verify our leg detection algorithm in several different scenarios to ensure safe operation. As future work, we propose to study an impedance control strategy in conjunction with model predictive control to ensure an optimal control strategy that keeps the robot within a strict distance and matched velocity to the patient. Sensing will be provided by embedded force/torque sensing in the robot arm in conjunction with the laser sensors. We also intend to add more functions to the MINA system like disinfection of a room, carrying an IV pole, and many others. We believe that such a robotic system will be useful in several hospital and nursing home scenarios to provide safe care, ensure nurse's safety, and reduce nurse burden.

\section*{Acknowledgment}
This work has been partially supported by National Science Foundation (NSF) grant IIP 1719031. 
This material is based upon work by the authors. Any opinions, findings, conclusions, and recommendations expressed in this paper are those of the authors and do not necessarily reflect the views of the NSF.

\end{document}